# Food Ingredients Recognition through Multi-label Learning

Rameez Ismail, Zhaorui Yuan

*Abstract*— The ability to recognize various food-items in a generic food plate is a key determinant for an automated diet assessment system. This study motivates the need for automated diet assessment and proposes a framework to achieve this. Within this framework, we focus on one of the core functionalities to visually recognize various ingredients. To this end, we employed a deep multi-label learning approach and evaluated several state-of-the-art neural networks for their ability to detect an arbitrary number of ingredients in a dish image. The models evaluated in this work follow a definite meta-structure, consisting of an encoder and a decoder component. Two distinct decoding schemes, one based on global average pooling and the other on attention mechanism, are evaluated and benchmarked. Whereas for encoding, several well-known architectures, including DenseNet, EfficientNet, MobileNet, Inception and Xception, were employed. We present promising preliminary results for deep learning-based ingredients detection, using a challenging dataset, Nutrition5K, and establish a strong baseline for future explorations.

## I. INTRODUCTION

What we eat and drink has a huge impact on our daily lives and our wellbeing. It is well established by now that a healthy and well-balanced diet is paramount to one's health. Daily diet varies considerably around the world, however, people in almost all regions of the world could benefit from rebalancing their diets by eating optimal amounts of various nutrients [1]. A suboptimal diet does not only carry risks for physical health but might reduce cognitive capabilities. Understandably, numerous studies, for example [2][3][4], implicate dietary factors in the cause and prevention of diseases such as, cancer, coronary heart disease, diabetes, birth defects, and cataracts. Similarly, findings from nutritional psychiatry indicate multitude of consequences and implications between what we eat and how we feel and ultimately behave [5]. On a population scale, eating habits and a broader diet pattern of a population is shown to be correlated with its health outcomes and longevity [6][7]. For example, a diet that adheres to traditional Mediterranean diet principles represents a healthy pattern and is positively associated with the longevity in Mediterranean blue-zones [8]. Although, such guidelines and general principles are quite useful, the individual metabolic responses varies substantially even if individuals are eating identical meals [9]. This calls for a personalized nutrition guidance approach that goes beyond general health recommendations. Personalized or precision guidance could additionally enable better management of various nutrient-related health conditions and diseases [10], for example, celiac disease, bowel syndrome, phenylketonuria, food allergies and diabetes. However, to make the proposition viable and to drive impact at scale, the guidance system must be automated.

A major difficulty in realizing an automated diet guidance is to capture the eating habits of a user accurately and effortlessly. Besides providing an introspection ability to the consumers, capturing dietary data from several participants is fundamental for understanding the diet and disease relationship. An accurate assessment of dietary intake enables the investigators to make progress in diet related studies by discovering patterns in context of nutritional epidemiology [11][12]. Diet assessment is usually performed using one of the three basic methods: meal recall, food diaries, or food frequency questionnaires. However, all these methods are based on self-reporting and therefore are time consuming, tiresome, and prone to misreporting errors. With recent advances in sensor technology and Machine Learning (ML) algorithms, automated food assessment has gained ground. Some of the technologies being explored in this direction include digital biomarkers and indigestible sensors [13][14], imaging sensors coupled with advanced AI/ML algorithms [15][16], as well as smart scales and eating utensils [17]. Digital cameras are readily available and can perform a detailed analysis on the food being consumed and the eating behaviors of the subjects but relies heavily on the visual recognition and the interpretation technology.

This work focusses on evaluating the performance of various state-of-the-art ML algorithms to detect ingredients or food-items present in a digital food image. Recognition of the actual ingredients is one of the two essential ML functions towards our envisioned food assessment system depicted in Fig. 1. The other is the portion estimation. When all the major ingredients and their portion size are identified, a descriptive nutrition log can be created by using a simple

Rameez Ismail and Zhaorui Yuan are with the team of embedded intelligence and analytics at Philips Research, High Tech Campus, 5656, Eindhoven, The Netherlands, Corresponding author: R. Ismail (rameez.ismail@philips.com). The research leading to these results has received funding from the European Union's ECSEL Joint Undertaking under grant agreement n° 826655 - project TEMPO.

lookup service that searches across various nutrition fact databases. There are several food databases in public domain which can be utilized for this purpose, such as USDA Food and Nutrient Database for Dietary Studies (FNDDS) and Dutch Food Composition Database (NEVO).

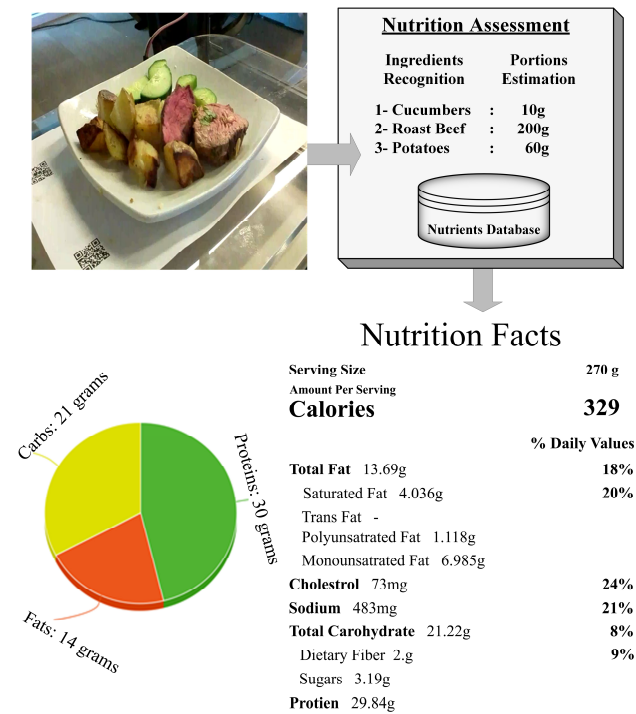

**Fig. 1.** The envisioned approach towards automated nutrition assessment. The assessment block consists of two ML functions, one for detecting the ingredients and the other to estimate the quantity, and a nutrients database look-up service.

In contrast to building separate ingredient recognition and portion estimation functions, some studies attempted an end-to-end nutrition estimation scheme [18], which estimates the total amount of macro-nutrients directly from the food images. Such end-to-end schemes, however, suffer from descriptive inadequacy. For example: actual food items composition and the respective portions remain undiscovered, which are important details for dietary research and for determining the eating habits of the users. Additionally, as the two-stage scheme depicted in Fig. 1 is more transparent, it is also expected to score better on the scale of user engagement and trustworthiness.

In summary, automated nutrition assessment has the potential to empower communities, on one hand, while enables investigators to make progress in nutritional epidemiology on the other hand. The main contributions of this work are: (a) motivate and investigate the problem of multi-item food recognition and proposes a multi-label learning framework to achieve this, and (b) evaluates and benchmark the proposed framework using various state-of-the-art deep learning modules and a challenging dataset, Nutrition5k [18], comprised of real-world food images with ingredient level annotations.

## II. Related Work

Automated food assessment through visual recognition has attracted a decent research interest in recent years. Existing efforts include deep learning based single dish recognition [19][20], contextualized food recognition (for example using GPS data which exploits the knowledge about location and data from the restaurants), multiple-food items detection [21][22] and real-time recognition [23]. This section mainly reviews the previous works on multiple food-item recognition using dish images as well as multi-label learning in general.

Some of the early approaches, towards multiple food-item recognition, relied on a separate candidate regions generation, using either simple circles or Deformable Parts Models (DPM), followed by features based classification of the candidate regions [24]. Such approaches employed various hand-crafted features such as color, texture, and Scale-invariant Feature Transform (SIFT) features for classifying the candidate regions. However, majority of the recent work is based on deep neural networks, because of its powerful feature representation ability, where such features are directly learnt from the data. For example, [25] proposed to use a convolution neural network GoogLeNet for labels prediction and employed the DeepLab [26] for semantic segmentation of the images. This pixel level image segmentation allows further analysis, such as estimation of the count and portion of the food constituents. More recently, PRENet [27] adopted a progressive training strategy to learn multi-scale features for a large-scale visual food recognition task. The approach utilizes a self-attention mechanism to contextualize the local features. These refined local features and a set of global features are then concatenated for the final classification task.

Both pure convolutional [20][25][18] and attention-based networks [27] have shown promising results for visual food recognition. Nevertheless, the latter can contextualize and dynamically prioritize the information. This suggests that the attention networks can extract much richer descriptions from the images compared to pure convolutional networks. However, it also implies that the learning process is comparatively difficult as the attention-based networks have more degrees of freedom. Moreover, since attention mechanism is a novel paradigm within deep learning, its potential is not yet fully explored for the task of ingredients and multiple food-item recognition.

Dish images in real-life usually contain multiple food items and ingredients, which makes it worthwhile to detect multiple labels independently for each input image. It also provides an added benefit that any food image can be inferred by the model, even if the actual dish is novel to the model, given that all its constituents were taught to the network during the training phase. The independent prediction of various labels against a single image is a more general classification problem, commonly known as multi-label classification or extreme classification. In such a regime of classification [28] [29], the goal is to predict the existence of a multitude of classes, thus forcing the model and training scheme to be efficient and scalable. The networks usually contain a stem and a head: the stem outputs a spatial embedding while the head transforms this spatial embedding into prediction logits. The most employed head is a Global Average Pooling (GAP) layer, which computes a scalar global embedding for each spatial embedding, followed by a fully connected read-out layer.

In case of multi-label learning, each neuron in the read-out layer is a binary classifier representing a specific class. Recently, several works proposed novel attention-based heads for multi-label classification. For example, [30] proposed an approach which leverages Transformer [31] decoders to query the existence of various class labels. Similarly, [32] introduced a class specific residual attention (CSRA) module that generates class-specific features, using a simple spatial attention score, and combines them with the global average pooling achieving state-of-the-art results. ML-Decoder [33] is yet another classification head based on transformer decoder architecture, which outperforms on various multi-label classification. One of the critical challenges for multilabel image classification is to learn the inter-label relationships and the dependencies.

The transformer-based methods employ attention mechanism with the goal to implicitly model the co-occurrence probability of the labels through relative weighing of the latent embeddings. Correspondingly, graph convolutional networks for multilabel learning [34] is another emerging theme that captures the complex inter-label associations by modelling them as graph nodes. In this work, we constrained our analysis to networks that follows a definite meta-structure shown in Fig. 2. The encoder block extracts the discriminative features from the image. Thus, projecting the image onto a latent space, while the decoder predicts the presence of various labels using the latent feature space.

III. METHODOLOGY

In this section, we first describe various deep learning modules used to build the classification networks benchmarked in this work. Then we explore the dataset, *nutrition5k,* used to train and evaluate the networks, followed by explanation of the loss function, utilized for the training, and the evaluation metrics.

*A. Multi-Label Classification Networks*
All classification networks benchmarked in this work are composed of an encoder and a decoder module, as depicted in Fig. 2.

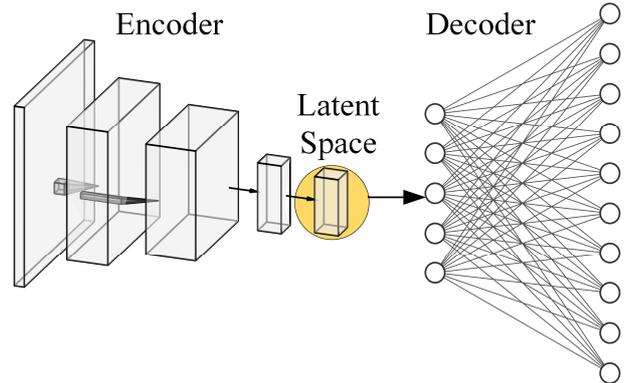

**Fig. 2.** The meta-structure used for the evaluated models.

The encoder performs features extraction, using the input image, while the decoder exploits these features to construct an ingredient breakdown sequence. The decoder can be as simple as a linear learnable projection that maps the latent space on to a layer of read-out neurons but can also have a complex construction. For example, a versatile attention block that exploits inter-label relationships through a querying mechanism or an autoregressive decoder that contextualizes its mapping based on its previous predictions. Next, we describe the various implementation of these blocks implemented and benchmarked in this work:

1) **Encoder**
   An encoder block can be modelled by any network that projects the image onto a latent space, $F \in \mathbb{R}^{H \times W \times D}$. For example, both convolutional neural networks, which progressively construct richer representations using convolution operations, and transformer networks, which transforms images into a small set of embedding tokens, are great candidates for the encoder block. In this work, however, we limited our analysis to convolutional networks for encoding. The employed encoding networks include, MobileNet [35][36], DenseNet [37], Inception network [38], Xception network [40] and EfficientNet [39]. These are some of the top-performing networks, when tasked to perform single label classification, evaluated on a large-scale visual recognition challenge ImageNet [41] dataset.

2) **Decoder**
   We used two distinct approaches to model the decoder block, the first approach employs a simple GAP based decoder while the second exploits an attention-based group-decoding scheme, ML-Decoder [33]. In a GAP based decoder, we first project the features, $F \in \mathbb{R}^{H \times W \times D}$, on to a single dimensional vector, $z \in \mathbb{R}^D$, by averaging over the spatial dimensions. Afterwards, a dense layer transforms the single dimensional vectors through a learnable linear projection into $K$ logits. $K$ is the number of total classes. The ML-Decoder is adapted from the transformer-decoder [31] with the goal to meet the computational demands for multi-label learning as the computational cost grows quadratically with the number of classes. The proposed modifications include the removal of a self-attention block, which reduces the quadratic dependency of the decoder in the number of query tokens to a linear one, and the introduction of a group decoding scheme. In a group-decoding scheme, a single query token is responsible for decoding multiple ingredients, thus limiting the required number of query tokens.

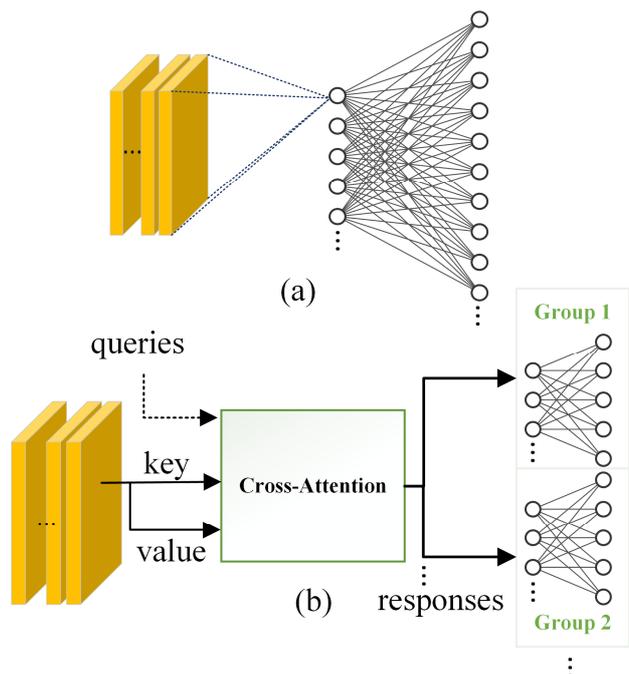

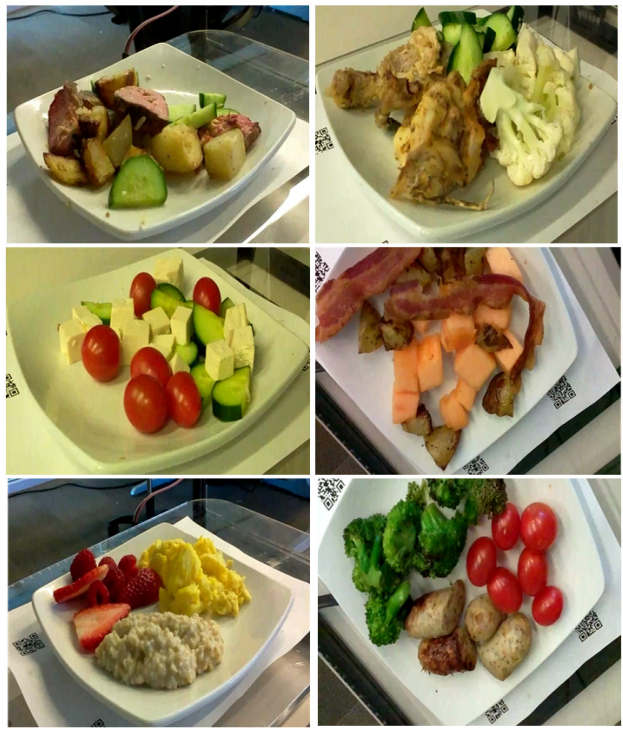

**Fig. 3.** The decoding mechanisms evaluated in this work: (a) the GAP based decoder averages the latent features before projecting them on to a read-out layer, and (b) the ML-Decoder that computes a response, linear combination of value vectors, against each external query vector. The responses are then projected on to the read-out layer through a group decoding scheme [33].

This strict decoupling of the roles enables the reuse of the latent space across various image recognition tasks. Fig. 3 provides a visual description of the two decoding schemes evaluated for the task of multi-label learning.

*B. Dataset*

We employed Nutrition5k [18] to train and evaluate various ML models explored in this work. Nutrition5k is a relatively diverse dataset of mixed food dishes with ingredients level annotations. The dataset contains 20k short videos generated from roughly 5000 unique dishes composed of over 250 different ingredients. The dataset also contains the portion estimates of each ingredients, which makes it possible to perform a supervised learning for portion estimation on top of ingredients recognition. The original dataset is collected using video cameras, mounted on the sides of a custom platform to capture each dish from various angles. A digital scale was embedded under the food plates to weigh the dish contents.

**Fig. 4.** Examples of dish images from the test dataset.

The dataset exploits an incremental scanning approach, where a plate is scanned at various time instances with growing cardinality of the ingredients. This resulted in a rich and diverse set of images with varying portion sizes, ingredients, and dish complexity. All incremental scans from a single dish image were organized into a single split to avoid any potential leak of information between the train and test split. Besides side angle videos, the dataset also exhibits a smaller subset of images collected from a top mounted RGB-D camera, which provides depth images from a top-down view. We constrained our analysis to RGB images only collected from the side-angle video cameras. The data is organized into a `train` and `test` subsets, following the original train-test splitting of the video files. The final dataset is obtained by extracting a single image from each video file. To this end, we simply extracted the first frame from each video and downsized it appropriately to be able to process through the neural networks.

This results in an image dataset comprising of around 15K training images and 2.5K test images, all of which are resized to 448x448 resolution in pixels. In Fig. 4, few examples of the dish images from the test split of the dataset are depicted.

*C. Loss Formulation*

Within multi-label learning, a common loss formulation is to consider the final read-out neurons as a series of mutually independent binary classifiers and compute their aggregated binary cross-entropy score. This aggregated score is then used as the minimization objective for the multi-label learning.

Given K labels, each neuron in the read-out layer outputs a score, $z^k$ where $k \in K$, that represents exclusively a single label. Each neuron is then independently activated by a sigmoid function $\sigma(z^k)$, which converts the logits into probability scores. Let's denote $y^k$ as the ground-truth for the k$^{th}$ class, the total classification loss $L^{total}$ is then obtained by aggregating a binary loss from all labels.

$$L^{total} = \sum_{k=1}^{K} L(\sigma(z^k), y^k) \quad (1)$$

A general form of a binary cross entropy loss per label, L, is given by:

$$L = -yL^+ - (1-y)L^- \quad (2)$$

Where, $L^+$ and $L^-$ are the are the positive and negative parts of the loss respectively, normally evaluated by $L^+ = log(p)$ and $L^- = log(1-p)$. These parts can be additionally weighted to asymmetrically focus more on the presence or absence of the label. A form of this asymmetric weighing, for multilabel learning, is proposed in [42], which introduces independent focusing parameters, $\gamma^+$ and $\gamma^-$ for positive and negative loss parts. This updates the computations as given by (3)

$$L^+ = (1-p)^{\gamma^+} log(p)$$
$$L^- = p^{\gamma^-} log(1-p) \quad (3)$$

We used the asymmetric loss, given by (2) and (3), for evaluating all models in this work. The focusing parameters are set to $\gamma^- = 5$ and $\gamma^+ = 0$; which effectively down weighs the loss contribution from easy negatives allowing the network to focus on harder samples as the training progresses.

*D. Evaluation Metrics*

We employed a mean Average Precision (mAP) metric to evaluate the performance of all ingredient recognition networks, which is a common practice in multi-label learning tasks. The average precision for a single prediction is computed using (4), a micro-average of this is then computed over all the predictions and samples to compute the final score.

$$AP = \sum_n (R_n - R_{(n-1)}) P_n \quad (4)$$

Where $R_n$ and $P_n$ are the recall and precisions at the n$^{th}$ thresholds. We used a total of 500 thresholds, equally distributed on the interval [0, 1], to calculate the individual recall and precision scores.

IV. RESULTS

Table 1 outlines the best mAP scores obtained from the several models evaluated in this work, along with the compute specifications of these models. The compute cost is parameterized by the number of atomic operations and the parameters in the models. The reported mean average precision score is obtained by evaluating the models on the test split of the dataset.

TABLE I
PERFORMANCE EVALUATION AND BENCHMARK

| Models | | Performance | Compute | |
|---|---|---|---|---|
| Encoder | Decoder | mAP % | Operations (GFLOPs) | Parameters (MParams) |
| DenseNet121 |  | 75.6 | 22.6 | 7.6 |
| DenseNet169 |  | 76.5 | 26.9 | 13.6 |
| DenseNet201 |  | 74.7 | 34.3 | 19.4 |
| MobileNetV1 |  | 72.4 | 4.5 | 3.8 |
| MobileNetV2 | GAP | 74.5 | **2.4** | **2.9** |
| EfficientNetB0 |  | 73.3 | 3.1 | 4.8 |
| EfficientNetB1 |  | 71.9 | 4.6 | 7.3 |
| EfficientNetB2 |  | 72.2 | 5.3 | 8.6 |
| EfficientNetB3 |  | 71.5 | 7.8 | 11.6 |
| EfficientNetB4 |  | 72.7 | 12.9 | 18.7 |
| Xception |  | **78.4** | 36.6 | 22.0 |
| InceptionV3 |  | 72.8 | 26.4 | 23.0 |
| MobileNetV2 |  | 68.0 | 3.4 | 9.3 |
| EfficientNetB0 | ML-Decoder | 73.4 | 4.2 | 11.1 |
| DenseNet169 |  | 67.9 | 28.0 | 19.9 |
| Xception |  | 70.0 | 38.0 | 28.5 |

In the first set of experiments, we explored a global average pooling (GAP) based decoding, with different classification neural networks acting as the encoder. In the second set, we selected four encoders to couple with the ML-Decoder, based on their performance in the previous experiments and the compute specifications. The resulting models are then trained with an identical configuration as the previous experiment.

Among the models with the GAP decoder, Xception network achieved the best mAP score of 78.4% but is also one of the most computationally expensive networks: it requires 36.6 billion multiply-add operations per image inference. Among smaller networks, the mobileNetV2 is exceptionally performant with 74.5% mAP while using only 2.4 billion operations per inference. Table 1 shows that not all computationally sophisticated models perform equally well. For example, the EfficientNetB1-B4 and the inceptionV3 models did not show improvements despite their high compute and memory complexity. Upon inspection of the training and validation accuracy curves, we concluded that that these larger models were overfitting the training data and that their performance is capped by the availability of data. However, not all larger models suffer equally from the over-fitting problem. For instance, DenseNet and Xception are comparatively bigger and yet are able generalize quite well on the test dataset. Furthermore, we observed that EfficientNet models require more careful calibration to reach their full potential. However, to create a fair benchmark, we did not attempt model-specific hyper-parameter tuning sweeps. All networks are trained using a standard training configuration as described in the appendix. All models with GAP decoder scored more than 70% mAP, showing the effectiveness to this simple decoding scheme.

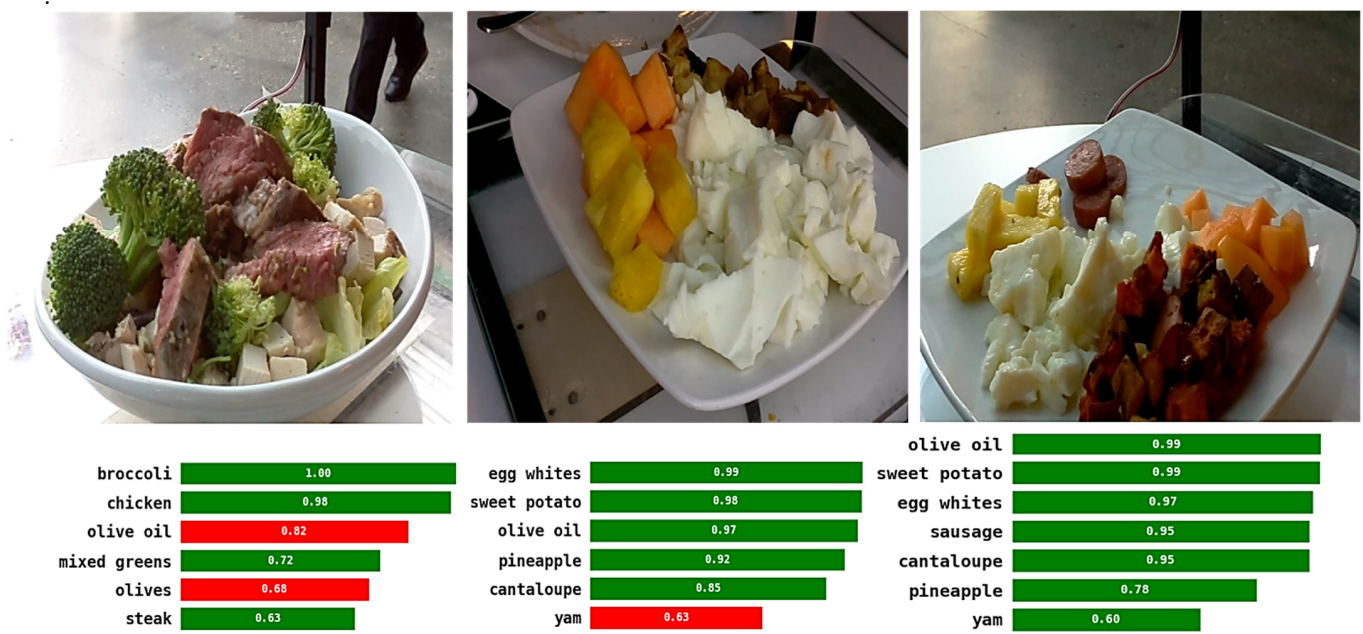

**Fig. 5.** Prediction results for the dish images selected from the test dataset for illustrative purpose. The detection confidence for each ingredient is shown. The green color represents a true positive detection while the red stands for the false positives.

The ML-Decoder did not perform well for the task of ingredients detection. As claimed in [33], the decoder is meant to be a drop-in replacement for GAP-based decoder in a multi-label learning setting. However, our results do not support this claim. Although, some networks perform equally good when coupled with ML-Decoder, this does not hold in general for all encoder models. For example, when coupled with EfficientNetB0, ML-Decoder performs slightly better compared to using the GAP decoder, while when coupled with other encoders the performance deteriorates. The large performance difference between the models with MobileNetV2 and EfficientNetV2 encoder highlights the weakness of the ML-Decoder, as both encoders are of similar compute sophistication. The root cause analysis of this issues and an ablation study of the decoder is currently considered for future work. Fig.5 demonstrate few test images, annotated with predictions from a trained model, composed of Denset121encoder and a GAP decoder.

## V. CONCLUSION

In this work, we presented a framework for automated diet assessment and demonstrated encouraging results for image-based ingredient recognition using deep learning. The Xception encoder, coupled with a global average-pooling based decoding scheme performs the best, with a mean average precision score of 78.4%. It therefore creates a strong baseline for future work. The attention-based decoding, contrary to what we conjectured, is unable to reliably extract the inter-label relationships and does not improve the overall performance. An ablation study of the attention-based decoder will be attempted further to presumably overcome its current limitations.

## APPENDIX

For all training jobs, we used Adam optimizer with an initial learning rate of $1e^{-3}$ following a learning schedule with linear warmup of 200 iterations and a cosine decay to $1e^{-6}$ afterwards. The batch size for all trainings was set to 32 and the models were trained for a maximum of 50 epochs. The encoder of each model is initialized with ImageNet weights. For the ML-Decoder we report outcomes from its default multi-label configuration [33]. Furthermore, we applied a simple data augmentation pipeline, which is composed of random horizonal and vertical flips, random image translations, and random crops with padding.

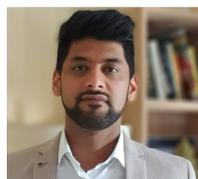

**Rameez Ismail** received master in robotics (2015) from Technical Univ. Dortmund, Germany, and Engineering Doctorate in systems design (2017) from Eindhoven Univ. of Technology, Netherlands. He started his career with a brief stint (2016-2018) at NXP Semiconductors and thereafter joined Philips Research as a scientist with the ambition to accelerate the digital transformation of healthcare. His research interests include Artificial Intelligence (AI), biomimetics systems and High-Performance Computing (HPC). Over the years, he successfully applied advanced AI research to various healthcare use cases within the scope of personal health and medical imaging.